\documentclass[journal]{IEEEtran}

\IEEEoverridecommandlockouts
\usepackage{cite}
\usepackage{amsmath,amssymb,amsfonts}
\usepackage{algorithmic}
\usepackage{graphicx}
\usepackage{textcomp}
\usepackage{xcolor}
\usepackage{comment}
\def\BibTeX{{\rm B\kern-.05em{\sc i\kern-.025em b}\kern-.08em
    T\kern-.1667em\lower.7ex\hbox{E}\kern-.125emX}}

\usepackage[utf8]{inputenc}
\usepackage{xr}
\usepackage{mathtools}
\usepackage{enumitem}
\usepackage{ragged2e}
\usepackage{authblk}

\usepackage{subcaption}

\begin{document}
\title{Autonomous Strike UAVs for Counterterrorism Missions: Challenges and Preliminary Solutions}

\author{Meshari Aljohani}
\author{Ravi Mukkamalai}
\author{Stephen Olariu}

\affil{Department of Computer Science, Old Dominion University\\ Norfolk, Virginia, 23529, USA}
\affil{\{maljo001, rmukkama,  solariu\}@odu.edu}

\maketitle

\abstract{Unmanned Aircraft Vehicles (UAVs) are becoming a crucial tool in modern warfare, primarily due to their cost-effectiveness, risk reduction, and ability to 
perform a wider range of 
activities. The use of autonomous UAVs to conduct strike missions against highly valuable targets is the focus of this research. Due to developments in 
ledger technology, smart contracts, and machine learning, such activities formerly carried out by professionals or remotely flown UAVs are now feasible. 
Our study provides the first in-depth analysis of challenges and preliminary solutions for successful implementation of an autonomous UAV mission. 
Specifically, we identify challenges that have to be overcome and propose possible technical solutions for the challenges identified. We also derive analytical 
expressions for the success probability of an autonomous UAV mission, and describe a machine learning model to train the UAV.}


\section{Introduction and motivation}

For several decades, the United States has employed remotely piloted Unmanned Aircraft Vehicles (UAV) across its military services \cite{dod-2017}. As several 
military analysts have pointed out, UAVs are attractive from both strategic and tactical standpoints because they are cheaper to deploy than crewed 
(i.e., manned) aircraft and they can carry out dangerous missions without risking human lives \cite{ahmed-2022}.

In addition, with the gradual introduction of more and more sophisticated UAVs, supported by advances in machine learning (ML), several new types of missions 
are within reach. 
These include cargo and resupply, air-to-air combat, close air support, communication relays, aerial refueling, search-and-rescue, and counter-terrorism 
missions \cite{hoehn-2022}. 
It is becoming evident that, due to increased technological sophistication and reduced size, UAVs are well-suited to carry out many types of missions that, 
until very recently, could 
only be performed successfully by crewed aircraft. Such considerations could enable the military to station UAVs closer to the front lines than crewed aircraft, 
potentially reducing the time to carry out time-sensitive missions.

As the US is withdrawing from conflicts around the world, the military will have to increasingly rely on UAVs 
for various missions including intelligence, surveillance, and acquisition of ground targets in counter-terrorism missions \cite{hoehn-2020,crs-2022}.

It is widely known that the U.S. Department of Defense (DOD) is developing several experimental concepts such as aircraft system-of-systems, swarming, and lethal 
autonomous weapons that explore new ways of employing future generation UAVs \cite{schneider-2017,andersen-2018,harrison-2021,hoehn-2022}.
Aligned with this effort, the main vision of this paper is to take UAVs to the next level of sophistication by enabling autonomous UAVs to carry 
out strike missions against entrenched high-value terrorists. 

While, in the past, such missions were carried out by Special Operations personnel and/or by remotely-piloted UAVs, recent advances in blockchain 
technologies, smart contracts, and ML have made it possible for these missions to be carried out successfully by autonomous UAVs. Our first main contribution 
is to identify the main challenges that have to be overcome to implement our vision; our second main contribution is to propose preliminary solutions to 
those challenges. To the best of our knowledge, ours is the first paper in the open literature available to us, that discusses the challenges inherent 
in making such missions feasible and the ways in which these challenges can be successfully overcome.

The remainder of the paper is structured as follows: Section \ref{sec:related-work} reviews related work. Following this, Section \ref{sec:background} provides 
the necessary background information. Section \ref{sec:scenario} introduces our working scenario and basic assumptions. Section \ref{sec:challenges} identifies some of the 
main challenges involved in enabling autonomous strike UAVs. Next, Section \ref{sec:tech} provides preliminary solutions to the challenges identified in Section 
\ref{sec:challenges}. Section \ref{sec:tasks}, given a collection of tasks $T_1,\ T_2,\ \cdots,\ T_n$ that have to be executed as part of the mission, evaluates 
the conditional probability of a successful completion of a future task, say $T_{i+1}$, given the completion status of the currently executed task, $T_i$. Further, 
Section \ref{sec:sensors} identifies some of the sensors that are provided in the UAV in support of its autonomous mission. Section \ref{sec:ML} offers the details 
of our ML framework in support of autonomous strike UAVs as well as a host of empirical evaluations. Before concluding, Section \ref{sec:discussion} 
delves into a discussion about advanced ML techniques and their relevance to our research. Finally, Section \ref{sec:concl} offers concluding remarks and 
maps out directions for future work.

\section{Related Work}\label{sec:related-work}
The rapid growth of UAV technology, particularly in military applications, has created a lot of attention and development in recent years. This section looks into the latest 
literature and research that has contributed to the increasing use of UAVs in modern military and intelligence operations.

The authors of \cite{kurunathan2023} conduct a comprehensive survey of the use of ML in the context of UAVs. It begins by addressing the growth of UAVs and 
how ML  might help them perform better. The survey then organizes the use of ML into four categories: perception and feature extraction, feature interpretation and 
regeneration, trajectory and mission planning, and aerodynamic control and operations. The survey describes several ML algorithms and strategies for each category, demonstrating 
how they are used to improve UAV operations such as image processing, object detection, autonomous navigation, and data transmission. It stresses the role of ML in 
improving UAV intelligence for activities such as environmental monitoring, surveillance, and communications. The paper also discusses the challenges involved in integrating ML 
with UAVs, such as processing restrictions, data management, and energy efficiency. It finishes with suggestions for future research initiatives, including the creation of 
more advanced ML models for UAVs operating in diverse and complicated situations. This comprehensive assessment emphasizes the expanding role of ML in the advancement of UAV technology.

The authors of \cite{ning2023} give a comprehensive review of the relationship between mobile edge computing (MEC), MML, and UAVs in the context of the Internet of Things (IoT). 
The study presents a thorough assessment of the most recent advances and uses of MEC and ML in UAV networks. It addresses the advantages and disadvantages of combining 
various technologies, focusing on their potential to improve the performance, efficiency, and capacities of UAV systems. The survey also looks at several scenarios and use 
cases, emphasizing MEC and ML's revolutionary impact on UAV operations in a variety of settings and applications. The paper concludes by discussing future research 
issues and expected advances in this growing field.

The authors of \cite{bai2023} present a detailed review of the use of Reinforcement Learning (RL) in Multi-UAV Wireless Networks (MUWN). It investigates numerous elements of RL 
for improving UAV operations, including data access, sensing, collection, resource allocation, edge computing, localization, trajectory planning, and network security. 
The paper examines the particular issues of implementing RL in UAV networks, including computing limits and changeable environmental variables. It finishes with suggestions 
for future research topics and the creation of sophisticated RL models to increase the efficiency and effectiveness of MUWNs.

The authors of \cite{tahir2019} provide a detailed examination of constructing a UAV system designed for anti-terrorist activities. It offers a thorough examination of numerous 
components, such as electrical systems, sensor systems, vision systems, ground control stations, propulsion systems, and structural systems. The study examines various 
UAV models, evaluating their capabilities using variables such as payload endurance, cost, and system components. It also recommends changes to the chosen UAV platform to 
improve its performance for anti-terrorist missions. The report concludes with a thorough evaluation of the effects of these adjustments on the UAV's performance, emphasizing 
the need for adapted architecture in UAV development for specialized missions.

The authors of \cite{matthew2021} investigate the application of AI-powered UAVs for security and surveillance, especially in challenging environments such as dense woods. 
It concentrates on the development of UAV systems that use cutting-edge technology such as laser-range detectors for exact location evaluation and path-finding, as well as 3D 
mapping capabilities for comprehensive environmental awareness. The research underlines the potential of using AI-powered UAVs to improve security measures. The use of 
convolutional neural networks and IoT frameworks is also covered, demonstrating how these technologies may transform environmental sensing, security monitoring, 
and search and rescue operations.

In their discussion on the employment of AI-powered UAVs for military purposes, the authors \cite{petrovski2022} categorize various UAV types according to factors including 
weight and flying characteristics. It suggests identifying vital military assets like trucks and artillery from UAV surveillance imagery in real-time using the YOLOv5 deep 
learning algorithm. The model was trained using a dataset of more than 10,000 tagged photos of military hardware, and it demonstrated good accuracy. AI-powered UAVs can offer 
improved battlefield awareness and intelligence when they are connected to military command and control networks. Important benefits including force multiplication and less 
risk to human life are highlighted in the research, but problems like false positives require development. Overall, it shows how automated detection and autonomous 
capabilities may be added to UAVs through deep learning and computer vision, eliminating the need for human operators in numerous dangerous military operations. 
The study highlights system interoperability and offers a technique for YOLOv5-based real-time important item recognition from UAV footage.

The authors in \cite{ahamed2022t} present a distributed blockchain-based platform for UAVs. Its main goal is to improve security and operating autonomy in an IoT setting. The proposed solution includes a special, secure, and light blockchain structure made just for UAV communication. This reduces the need for computing power and storage space while still providing privacy and security benefits. To make sure the autonomous network is reliable, a reputation-based consensus system is created. Different kinds of transactions are set up for different types of data access. The platform protects drone-based apps from possible vulnerability by using simple cryptography, new transaction and block structures, and a consensus method similar to Delegated Proof of Stake (DPoS) along with a reputation rating system. Performance reviews of the system show that it is good at lowering latency, speeding up data flow, and making security stronger against harmful threats. 

\section{Background} \label{sec:background}

UAVs are a big step forward in technology that can be used in a growing number of different areas. Smart contracts and blockchain technology could be used to manage the tasks of UAVs. This is a new and possible concept, especially for military missions where safety, autonomy, and reliability are very important. Using a private blockchain for this has many benefits, such as better protection, more limited access, and faster transaction times compared to public blockchains. The main characteristics of private blockchains are their high transaction processing pace and limited access, which only permits a small number of approved users to interact with the network. Compared to public blockchains, which have slower transaction rates because they require network-wide consensus, frequently via Proof-of-Work processes, this results in faster consensus times and more transactions completed per second. Compared to public blockchains where data is immutable and modifications need agreement across all subsequent blocks, private blockchains offer increased data privacy since changes may be made quickly after consensus is reached across all nodes \cite{yang2020}. 

\subsection{Smart Contracts}
SC technology is directly related to blockchain, the underlying platform that allows these contracts to function with the highest security and transparency. SCs, or 
self-executing programs embedded in blockchain, are transforming digital interactions. Blockchain acts as a decentralized ledger, recording all transactions across a 
network of computers. This architecture not only ensures SCs' immutability and traceability but also eliminates the need for a central authority or middleman, 
resulting in a more direct and transparent form of engagement and transaction execution \cite{aljohani2023}. The combination of SCs and blockchain technology enables a new era of automated, 
safe, and effective digital transactions, offering a wide range of opportunities across different sectors \cite{Smartcon82, Whatares16}.

\subsection{Real-world Example of Using Smart Contracts}

\begin{enumerate}

 \item \textbf{Supply Chain and Logistics}: Takes into consideration a situation where supply chain logistics, particularly in remote or difficult-to-reach places, 
are handled by UAVs. Logistics tasks including confirming delivery completion, automating payments after successful deliveries, and guaranteeing adherence to predetermined 
delivery guidelines can all be managed by smart contracts on their own. This guarantees supply chain accountability and transparency while also streamlining the logistical process.

\item \textbf{Agricultural Monitoring}: UAVs with SC capabilities can be utilized for crop management and monitoring in precision agriculture. When specific criteria 
are satisfied, SCs can evaluate data gathered by UAVs, such as crop health and soil humidity, and carry out pre-programmed actions, such as releasing water or fertilizers. 
By automating and streamlining agricultural processes, this may boost productivity and efficiency.

\item \textbf{Disaster Response and Relief Operations}: UAVs can be important in disaster response and relief operations. Based on real-time data analysis, SCs can be 
designed to allocate resources and control UAVs for specific purposes, such as determining which locations require the greatest assistance or keeping an eye on the 
distribution of relief goods. Disaster response activities can be made much more efficient and timely with the help of this program.

\item \textbf{Environmental Monitoring}: UAVs equipped with SCs can be used to keep an eye on protected regions as part of environmental protection programs. For example, 
UAVs can patrol zones on their own and, when they come across illicit activities like poaching or logging, they can carry out pre-programmed actions specified 
by the SC, such as notifying the authorities or taking pictures of the activity.
\end{enumerate}

\section{Working scenario -- A high-level description}\label{sec:scenario}

It has been recognized \cite{tahir2019} that UAVs have tremendous potential for air-to-ground strike missions. A strike UAV has the capability to launch weapons such as 
{\em precision-guided missiles} against a ground target. While the state of the art in air-to-ground strike missions is that there is always a man in 
the loop, in the sense that the UAV is piloted remotely, the vision of our work is to leverage the latest technology to enable fully autonomous strike UAVs.

With this in mind, throughout this paper, we assume that a UAV is deployed in support of a strike mission 
involving a high-value terrorist target in a foreign country. Such missions may well operate in ``contested territory’’ in which terrorist forces have an 
active presence. By their nature, these missions are top secret and do not rely on intelligence collected from foreign state actors. In fact, the mission 
may be deployed without the specific approval of foreign state actors. 

Given the context of the mission we are contemplating, we assume that the targeted terrorist organization does not have the wherewithal to take 
out or jam US communication satellites and, consequently, we rely on satellite-to-UAV communications for the duration of the mission.

We further assume that the UAV carries, as part of its payload, standard on-board sensory equipment, including a gyroscope (or inertial navigation system), 
electro-optical cameras, and infrared (IR) cameras for use at night, as well as synthetic aperture radar (SAR). SAR is a form of radar used to create 
two- or three-dimensional reconstructions of objects, such as landscapes. SAR uses the motion of the radar antenna over a target 
region to provide finer spatial resolution than conventional radars. Such missions must avoid civilian casualties. In fact, we assume that the mission will 
be aborted if civilians are close to the intended target. In this regard, night missions are safer to execute, because civilians (especially children) are 
very unlikely to be present, but they require far more sophistication in terms of localization and imagery processing.

\begin{figure}[!ht]
    \vspace{8pt}
    \centering
    \includegraphics[scale=0.5]{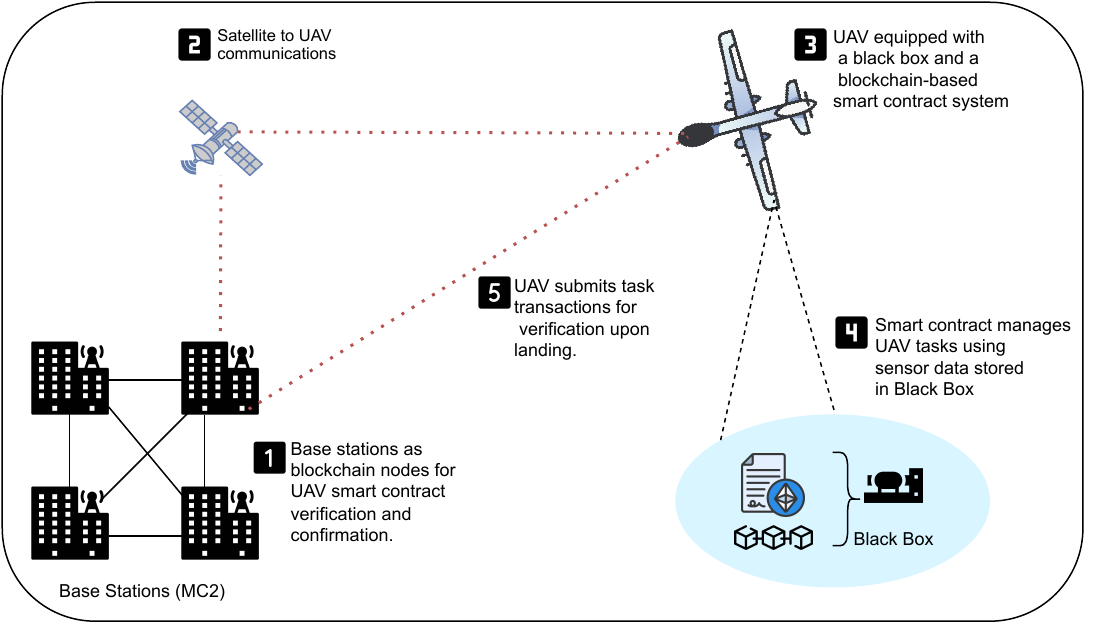}
    \caption{\em A comprehensive overview of our working scenario.} 
    \label{fig:blackbox}
\end{figure}

Figure \ref{fig:blackbox} provides a comprehensive overview of our working scenario. Here, we see a network of systems working together to achieve a targeted 
mission. The systems consist of a base station referred to as a Mission Control Center (MC2), equipped with blockchain systems and smart contracts, alongside a satellite communication system. Additionally, each UAV is equipped with an onboard Black-box (BBX) and an integrated blockchain system and smart contract within the UAV network. This integration demonstrates how blockchain technology is used in UAV networks to improve data integrity, operational autonomy, and security in addition to on the ground \cite{ahamed2022t, santos2021}. While this paper assumes a single UAV, the proposed system is equally applicable to multiple coordinated UAVs. The integration of a private blockchain allows the use of smart contracts for task management, ensuring that once the UAV returns, it reconnects with MC2 to transmit transaction data across the network for final validation.

\vspace{-10pt}
\unskip
\vspace{9pt}

Preparing a mission of the type we have in mind, requires the UAV to execute several {\em training runs} both day-time and night-time, each intended to 
evaluate the sequence of tasks (see Section \ref{sec:tasks}) that, collectively, make up the mission. The data collected in each such training run is 
carefully analyzed by human experts back at the MC2 to establish the conditional success probability of a future task given the status of the current one. 
Data from successive runs are aggregated by {\em human experts} at the MC2 and used to train an ML model. Specifically, human experts analyze the 
flight information stored in a tamper-proof on-board BBX, e.g., collected UAV imagery along with maneuvers performed by the 
UAV in response to sensory information. As a result, human experts can assign conditional probabilities to
individual tasks based on the successful or partly failed status of the previous task in the sequence. 
As already mentioned, strict conditions for avoiding civilian casualties are stipulated and encoded as part of the SC that is to oversee the mission.

In our vision, once the UAV is trained, it is ready to carry out the mission autonomously, without the need for continuous communication with the MC2. 
This absence of communication is crucial for security purposes, as it prevents unauthorized access and tampering with the data or UAV operations. 
We note that the on-board BBX serves as a reliable and secure storage system for the gathered information, protecting it from potential 
breaches or corruption. 
%

\section{Challenges}\label{sec:challenges}

In order to make the vision of autonomous strike UAV a reality, several technical challenges must be overcome. These challenges are listed in this section. 
Preliminary solutions to the listed challenges are detailed in Section \ref{sec:tech}.

The challenges we have in mind include, but are not limited to the following:
\begin{itemize}
\item  Accurate UAV localization in space and time. As we discuss in Section \ref{sec:tech}, accurate 3-D UAV localization, in all weather conditions, is an ongoing effort of 
great interest to both the research and the user community. The challenge of locating UAVs in time contains, as a sub-challenge, the time synchronization of the UAV 
and the MC2. While initially the UAV and the MC2 are assumed to be synchronized, due to clock drift, synchronization may be lost, and periodic re-synchronization 
becomes necessary \cite{olariu-2004,ashraf-2005}. This re-synchronization may not be done directly with the MC2 in case of non-line-of-sight operations and 
has to be performed using the UAV satellite link. Since we assume unimpeded UAV-satellite communication, time synchronization is possible as long as the UAV has a 
link to one of the military satellites. It is worth noting that time synchronization involves passive listening and does not require active UAV communications 
with the satellite;
\item  Enabling secure communication between the MC2 and the UAV. In our philosophy such communications occur sporadically to minimize the likelihood of the UAV being
discovered. However, these sort of communications are vital when, for example, the MC2 wants to inform the UAV about the mission being aborted. Similarly, under specific conditions,
when it becomes clear that the mission cannot be successful (e.g. children are playing in close proximity to the target), the UAV will seek permission to abort the mission. 
In a Line-of-Sight scenario, these communications occur directly between the UAV and MC2. Otherwise, the communications will be between the UAV and one of the satellites 
in the supporting satellite constellation. In either case, such communications may involve time-dependent frequency hopping and, as such, require tight time synchronization 
between the two parties \cite{olariu-2004a};
\item  Tamper-proof collection and storage of accurate in-flight sensory data and the various maneuvers the UAV performs in response to the received sensory input. 
Indeed, reliable, untampered sensor data and the corresponding UAV responses collected during {\em training} missions are 
absolutely necessary to train the ML model. Such data is also essential for auditability 
purposes, especially if the mission is aborted. Finally, the provenance of each piece of sensory data and UAV response will have to be recorded and ascertained;
\item  Identifying the target and confirming that the target is clear of civilians. A fundamental requirement of a successful mission is to avoid civilian casualties;  
\item  Allowing dynamic changes of mission parameters. This presupposes that some form of reliable communication, either direct or indirect via a satellite, has been established 
between the MC2 and the UAV; 
\item BBX tamper-resistance: if downed or captured, the UAV must blank/destroy its BBX. 
\end{itemize}

\section{Technical details: addressing the challenges}\label{sec:tech}

The main goal of this section is to outline our preliminary solutions to the challenges identified in Section \ref{sec:challenges}.
Our solutions are, necessarily, sketchy but should give the reader a sense of what the technical solutions involve.

\subsection{Accurate UAV localization}\label{subsec:loc}

UAV localization is one of the fundamental challenges in autonomous navigation and has received extensive attention in the scholarly literature 
\cite{abdelsalam-2012,fox-1999,pandey-2018,zhao-2018,li-2020,espinosa-2022,youssef-2022}. Since, as already mentioned, satellite (e.g. GPS) communications are available 
to the mission, we contemplate using GPS for both synchronization (to within 100 nano-seconds) with the MC2 and UAV localization \cite{ahmed-2022,youssef-2022}.

\subsection{Enabling secure communication between the MC2 and the UAV}\label{subsec:comm}

One of the fundamental tasks that have to be performed as part of a successful mission is communication with the MC2. Depending on the specifics of the mission, 
two types of communication may be required. For missions deployed within about 50 miles of the MC2, line-of-sight communications may be used. For missions 
beyond 50 miles from the home base, non-line-of-sight (NLoS) communications are required. 
In this paper, we assume NLoS control communications between the MC2 and the UAV, forwarded through one of the several satellite constellations established, 
for similar purposes, by the US military \cite{hoehn-2022}. Recall that we assumed that the targeted terrorist organization does not have the capability to take
out or jam US communication satellites and, consequently, we rely on satellite-to-UAV communications for the duration of the mission.
If, however, the communication link to the MC2 is lost, the UAV is programmed to return to its base, and the mission is aborted.

\subsection{Tamper-proof collection and storage of flight data}\label{susec:bb}
In our vision, the workhorse of the mission is the on-board tamper-proof BBX.
\begin{itemize}
\item The BBX implements, for the duration of the current flight, the functionality of an {\em append-only ledger};
\item The BBX records and stores every piece of sensory data collected by the UAV's sensors, along with a time stamp and provenance information;
\item The BBX also records, for ML training and auditing purposes, the action(s) taken by the UAV in response to each sensory input. This will allow the human experts at the MC2
to evaluate the successful completion of various intermediate tasks making up the mission.
\end{itemize}

In the light of the above, it is clear that the BBX serves as an on-board command and control center. Pre-flight, the BBX is loaded with the individual tasks that 
make up the mission, 
with the aggregated conditional probabilities discussed in Section \ref{sec:tasks}, 
confirmed by human experts at MC2.
The integration of advanced BBX and append-only ledger (i.e., blockchain) technology creates a powerful and secure system for data collection and mission 
management. In our system, the BBX functions as an essential safeguard for the information collected by the UAV, while at the same time, 
ensuring integrity, traceability, transparency, security, and auditability.
Each data entry or transaction can be traced back to its origin, making it easier to verify its authenticity and identify any potential issues. 

\subsection{Identifying and confirming the target}\label{subsec:target}

As already mentioned, the missions we contemplate involve a number of training runs whose stated goal, among others, is to locate and identify the 
target (which, to fix ideas,  we assume the target to be an isolated building). The location of the building is acquired through day-time 
training missions and 
confirmed by a human expert at the MC2. The same is repeated during subsequent night-time training runs, where the target is 
confirmed using IR and SAR imagery. As already mentioned, the mission is aborted if civilians are identified close to the target. However, since we envision 
a night-time execution of the strike, the presence of civilians in close proximity to the target is a very unlikely event.

\subsection{Allowing dynamic mission changes}\label{subsec:changes}

It is essential for the MC2 to be able to order aborting a mission in progress. This is accomplished by sending a specific encoded message to the UAV 
using the satellite communication channel discussed above. Such an order will have to be confirmed by the UAV using a different communication channel, such as a different 
satellite in the constellation.

\subsection{Erasing BBX content upon capture}

The main idea is to prevent the adversary from identifying the mission parameters and the flight data.
There are several possible techniques for meeting this challenge \cite{ashraf-2005}. The standard procedure is to zero out the contents of the BBX. Yet another, 
more sophisticated, approach is to automatically generate fake data. The fake data can be generated beforehand and would replace the actual content of the BBX, with the intention to
mislead the enemy.

\section{Evaluating the probability of mission success}\label{sec:tasks}

We take the view that the {\em mission} that the autonomous UAV is tasked with, involves performing a {\em sequence} of tasks $T_1, T_2, \cdots, T_n, \cdots$. Let the sequence of 
tasks performed by a given UAV be observed and evaluated by the on-board SC and their outcome recorded by the on-board tamper-proof, append-only ledger, implemented 
as a standard on-board BBX.
While, in this work, we assume that the tasks are atomic and indivisible, in reality, each task could be an amalgam of other, simpler tasks. 
An example would be the localization task, one of the fundamental tasks to be performed by the UAV. This task typically involves a Markovian sequence of other tasks, i
each processing sensor readings and inertial system input \cite{fox-1999}.
In this context, a task is considered either ``successfully completed'' (``successful'', for short), or ``partially completed'' (``incomplete'', for short), 
depending on whether or not certain task-specific performance parameters are met. 

We assume that just prior to being deployed, the UAV's
on-board BBX  was loaded with a SC cognizant of the tasks to be performed and of the conditions under which the mission cannot be completed successfully
and should be aborted. We assume that such conditions are expressed in terms of the {\em unconditional} probability of a task being completed successfully. 
Since each task in the sequence of tasks contributes to the success of the mission if the probability of success of any task is below a 
mission-specific threshold, the SC will inform the MC2 and seek permission to abort the mission and return to base. 

In the light of the above, in the remainder of this work, we concern ourselves with reasoning about the outcome of a future task, given the observed sequence 
of completed tasks. As already mentioned, the goal is to advise the MC2 on the likelihood of mission failure as a result of improper completion of 
individual tasks.

For a positive integer $k,\  (k \geq 1)$, let $A_k$ denote the event that task $T_k$ has been completed successfully. Our basic assumption is that the following relation holds true
\begin{equation}\label{markov}
\Pr[A_{k+1} | A_k \cap A_{k-1}\cap \cdots \cap A_2 \cap A_1] = \Pr[A_{k+1} | A_k].
\end{equation}
Equation (\ref{markov}) states that the outcome of the $(k+1)$-th task, in terms of being successful or incomplete, only depends on the 
completion status of the previous task, namely task $T_k$, and not on the status of earlier tasks \cite{lindquist-1978,klotz-1973}. 

Recall that the SC associated with the mission stipulates the conditions that have to be met for a mission to be aborted.
In our derivation, $p_1= \Pr[A_1]$, the probability that the first task is successful, plays a distinguished role. A good example of the first task
to be performed as part of the overall mission is UAV localization. If the UAV is deployed from an aircraft, say, a C-130 transport aircraft, then 
with a high probability, the initial localization task will be successful. Otherwise, it might not. With this in mind, in Subsection \ref{sec:known} $p_1$ 
will be assumed to be known, while in Subsection \ref{sec:p1-unknown} $p_1$ will be assumed to be unknown.
We begin by introducing notation:
\begin{itemize}
\item Let $\Pr[A_{k+1} | A_k]$ be the conditional probability that the $(k+1)$-th task is successful given that the $k$-th task was successful; 
\item Let $\Pr[A_{k+1} | {\overline A_k}]$ be the conditional probability that the $(k+1)$-th task is successful given that the $k$-th task was incomplete; 
\item Let $\Pr[\overline A_{k+1} | A_k]$ be the conditional probability that the $(k+1)$-th task is incomplete given that the $k$-th task was successful;
\item Let $\Pr[{\overline A_{k+1}} | {\overline A_k}]$ be the conditional probability that the $(k+1)$-th task is incomplete given that the $k$-th task was incomplete. 
\end{itemize}
\noindent
We use $\tau_k(s,s),\ \tau_k(u,s),\ \tau_k(s,u),\ \tau_k(u,u)$, as shortcuts for \\ $\Pr[A_{k+1} | A_k],\ \Pr[A_{k+1} | {\overline A_k}],\ \Pr[\overline A_{k+1} | A_k],\
\Pr[\overline A_{k+1} | {\overline A_k}]$.\\
While these conditional probabilities are, in general, functions of $k$, we assume that they are all {\em time-independent}, and will drop 
any reference to $k$, writing $\tau(s,s),\ \tau(u,s),\ \tau(s,u),\ \tau(u,u)$. 
\noindent
We assume that these conditional probabilities are known for the specific mission in support of which the UAV is being deployed. Indeed, the various probabilities
can be learned during the training runs and are loaded into the BBX 
at deployment time. Certainly, the SC knows these conditional probabilities as well.
\subsection{When the probability $p_1$ is known}\label{sec:known}
In this subsection, assuming a known value for $p_1$ as well as the conditional probabilities defined above, we are turning our attention to the task
of finding the {\em unconditional} probability of
the event that the $n$-th task is successful, for $n \geq 2$. Using the Law of Total Probability we write
\begin{eqnarray}\label{ltp}
\Pr[A_n] &=& \Pr[A_n | A_{n-1}] \Pr[A_{n-1}] + \Pr[A_n | {\overline A_{n-1}}] \Pr[\overline A_{n-1}] \nonumber \\
         &=& \tau(s,s) \Pr[A_{n-1}] + \tau(u,s) \Pr[\overline A_{n-1}] \nonumber \\
         &=& \tau(s,s) \Pr[A_{n-1}] + \left [1 - \tau(u,u) \right ] \left [ 1- \Pr[A_{n-1}] \right ] \nonumber \\ 
         &=& \Pr[A_{n-1}] \left [ \tau(s,s) + \tau(u,u) -1\right ] + \left [ 1 - \tau(u,u) \right ] \nonumber \\
         &=& \lambda \Pr[A_{n-1}] + \mu,
\end{eqnarray}
where $\lambda = \tau(s,s) + \tau(u,u) -1$ and $\mu = 1 - \tau(u,u)$.
For further reference, we note that 
\begin{equation}\label{ss}
\tau(s,s) = \lambda + \mu.
\end{equation}
\noindent
In order to avoid trivialities, we assume that $\lambda$ is different from  $-1,\ 0$, and $1$. 
Indeed, $\lambda =-1$ implies $\tau(s,s) + \tau(u,u)  =0$ which, in turn,
implies 
that $\tau(s,s) = \tau(u,u) =0$. Similarly, $\lambda =1$ implies $\tau(s,s) + \tau(u,u) =2$ which, in turn,
implies 
$\tau(s,s) = \tau(u,u) =1.$ Finally if $\lambda = 0$ then $\Pr[A_n]= \mu$ for all $n \geq 2$. 
In such a case, however, the sequence of task outcomes is rather trivial. 
Thus, from now on, we assume $0 < | \lambda | <1$. Now, (\ref{ltp}) and $\Pr[A_1] = p_1$, combined, yield
the following recurrence: 
\begin{equation}\label{pan}
\Pr[A_n]  = \left\{
                  \begin{array}{l l}
                  p_1 & \mbox{for $n=1$;} \\
                  \lambda \Pr[A_{n-1}] + \mu & \mbox{for $n \geq 2$.}
                  \end {array}
                  \right.
\end{equation}
\vspace*{1mm}
\noindent
A simple telescoping argument affords us the following closed form for $\Pr[A_n],\ (n \geq 1)$:
\begin{equation}\label{pan-closed}
\Pr[A_n]  = \left [ p_1 - \frac{\mu}{1- \lambda} \right ] \lambda^{n-1} + \frac{\mu}{ 1 - \lambda}.
\end{equation}
\subsection{When $p_1$ is unknown}\label{sec:p1-unknown}
We now assume that $p_1$ is merely a parameter and that we have no knowledge about whether or not $A_1$ will be successful. 
In this case, we will be evaluating the following conditional probabilities:
\begin{itemize}
\item $\Pr[A_{k+1} | A_1]$ -- the conditional probability that the $(k+1)$-th task is successful given that the first task is successful.
We use $\rho_k(s,s)$ as a shortcut for $\Pr[A_{k+1} | A_1]$;
\item $\Pr[A_{k+1} | {\overline A_1}]$ -- the conditional probability that the $(k+1)$-th task is successful given that the first task is incomplete.
We use $\rho_k(u,s)$ as a shortcut for $\Pr[A_{k+1} | {\overline A_1}]$;
\item $\Pr[\overline A_{k+1} | A_1]$ -- the conditional probability that the $(k+1)$-th task is incomplete given that the first task was successful.
We use $\rho_k(s,u)$ as a shortcut for $\Pr[\overline A_{k+1} | A_1]$.
\item $\Pr[{\overline A_{k+1}} | {\overline A_1}]$ -- the conditional probability that the $(k+1)$-th task is incomplete given that the first task was also incomplete.       
We use $\rho_k(u,u)$ as a shortcut for $\Pr[\overline A_{k+1} | {\overline A_1}]$;
\end{itemize}
We note that the semantics of the conditional probabilities just defined do not allow us to drop the subscript $k$ from 
$\rho_k(s,s),\ \rho_k(u,s),\ \rho_k(s,u)$ and $\rho_k(u,u)$. 
We begin by evaluating $\rho_k(s,s)$.
\begin{eqnarray}\label{ross}
&& \rho_k(s,s) = \Pr[A_{k+1}|A_1] = \Pr[A_{k+1} \cap \left ( A_k \cup \overline A_{k} \right )|A_1] \nonumber \\
            &=& \Pr[A_{k+1} \cap A_k |A_1] + \Pr[A_{k+1} \cap {\overline A_k} |A_1] \nonumber \\
            &=& \Pr[A_{k+1} | A_k \cap A_1] \Pr[A_k | A_1] \nonumber \\
            &+& \Pr[A_{k+1} | {\overline A_k} \cap A_1] \Pr[{\overline A_k} | A_1].
\end{eqnarray}
Let us note that
\begin{itemize}
\item By (\ref{markov}), $\Pr[A_{k+1} | A_k \cap A_1] = \Pr[A_{k+1} | A_k] = \tau(s,s)$;
\item $\Pr[A_k | A_1] = \rho_{k-1}(s,s)$;
\item Similarly, $\Pr[A_{k+1} | {\overline A_k} \cap A_1] = \Pr[A_{k+1} | {\overline A_k}] = \tau(u,s) = 1 - \tau(u,u)$;
\item $\Pr[{\overline A_k} | A_1] = 1 - \Pr[A_k | A_1] = 1- \rho_{k-1}(s,s)$.
\end{itemize}
On replacing the expressions above into (\ref{ross}), we write
\begin{eqnarray}\label{ross-new}
\rho_k(s,s) &=& \tau(s,s) \rho_{k-1}(s,s) + \left [ 1 - \tau(u,u) \right ] \left [ 1- \rho_{k-1}(s,s) \right ] \nonumber \\
            &=& \rho_{k-1}(s,s) \left [ \tau(s,s) + \tau(u,u) -1 \right ]  + 1 - \tau(u,u) \nonumber \\
            &=& \lambda \rho_{k-1}(s,s) + \mu,
\end{eqnarray}
where $\lambda$ and $\mu$ were defined in Subsection \ref{sec:known}. As before, we assume $0 < | \lambda | <1$. 
Now, noticing that $\rho_1(s,s) = \tau(s,s)$, we obtain the following recurrence
for $\Pr[A_{k+1}|A_1],\ (k \geq 1)$:
\begin{equation}\label{rho-rec}
\rho_k(s,s)  = \left\{
                  \begin{array}{l l}
                  \lambda+\mu & \mbox{for $k=1$;} \\
                  \lambda \rho_{k-1}(s,s) + \mu  & \mbox{for $k \geq 2$.}
                  \end {array}
                  \right.
\end{equation}
\vspace*{1mm}
\noindent
Simple manipulations yield the following closed form for $\rho_k(s,s)$
for $k \geq 1$:
\begin{eqnarray}\label{rho-closed-0}
\rho_k(s,s) &=& \left [ \tau(s,s) - \frac{\mu}{1- \lambda} \right ] \lambda^{k-1} + \frac{\mu}{ 1 - \lambda} \nonumber \\
            &=& \left [ \lambda + \mu - \frac{\mu}{1 -\lambda} \right ] \lambda^{k-1} + \frac{\mu}{ 1 - \lambda} \nonumber \\
            &=& \frac{1 - \lambda - \mu}{1 - \lambda} \lambda^{k} + \frac{\mu}{ 1 - \lambda} \nonumber \\
            &=& \left ( 1 - \frac{\mu}{1 - \lambda} \right ) \lambda^{k} + \frac{\mu}{ 1 - \lambda},
\end{eqnarray}
which is the desired closed form for $\rho_k(s,s)$.
%
%
Next, we turn our attention to the task of evaluating $\rho_k(u,s)$.  We notice that
\begin{equation}\label{rho1us}
\rho_1(u,s) = \Pr[A_2 | \overline A_1] = \tau(u,s) = 1 - \tau(u,u) = \mu.
\end{equation}
\noindent
Further, for $k \geq 2$, we write
\begin{eqnarray}\label{rhokus}
\rho_k(u,s) &=& \Pr[ A_{k+1} | \overline A_1 ] = \Pr[ A_{k+1} \cap \left ( A_k \cup \overline A_k \right ) | \overline A_1 ]  \nonumber \\
            &=& \Pr[ A_{k+1} \cap A_k | \overline A_1 ] + \Pr[ A_{k+1} \cap \overline A_k | \overline A_1 ] \nonumber \\
            &=& \tau(s,s) \rho_{k-1}(u,s) + \left [ 1 - \tau(u,u) \right ] \left [ 1 - \rho_{k-1}(u,s) \right ] \nonumber \\
            &=& \left [ \tau(s,s) + \tau(u,u) -1 \right ] \rho_{k-1}(u,s) + 1-\tau(u,u) \nonumber \\
            &=& \lambda \rho_{k-1}(u,s) + \mu.
\end{eqnarray}
Notice that (\ref{rho1us}) and (\ref{rhokus}) lead quite naturally to the following recurrence describing the behavior of $\rho_k(u,s)$.
\begin{equation}\label{rho-rec-1}
\rho_k(u,s)  = \left\{
                  \begin{array}{l l}
                  \mu & \mbox{for $k=1$;} \\
                  \lambda \rho_{k-1}(u,s) + \mu  & \mbox{for $k \geq 2$,}
                  \end {array}
                  \right.
\end{equation}
from where we obtain the following closed form for $\rho_k(u,s)$ for $k \geq 1$:
\begin{equation}\label{rho-closed-1}
\rho_k(u,s) =\frac{\mu}{1- \lambda} \left [ 1 - \lambda^{k} \right ].
\end{equation}
Finally, we turn our attention to computing the {\em unconditional} probability of $A_n$. Conditioning on $A_1$, we write
\begin{eqnarray}\label{ltp-1}
\Pr[A_n] &=& \Pr[A_n | A_{1}] \Pr[A_{1}] + \Pr[A_n | {\overline A_{1}}] \Pr[\overline A_{1}] \nonumber \\
         &=& p_1 \rho_{n-1}(s,s) + (1-p_1) \rho_{n-1}(u,s) \nonumber \\
         &=& \rho_{n-1}(u,s) + p_1 \left [ \rho_{n-1}(s,s) - \rho_{n-1}(u,s) \right ] \nonumber \\
         &=& \left ( p_1 - \frac{\mu}{1 - \lambda} \right ) \lambda ^{n-1} + \frac{\mu}{1 - \lambda},
\end{eqnarray}
which, amazingly,  matches the expression of $\Pr[A_n]$ derived in (\ref{pan-closed}).
Equations (5) and (14) provide us with the probability of a mission of $n$ tasks being successful, given the relationship between adjacent tasks. This is very helpful in designing a UAV mission. Obviously, as the number of tasks in a mission increases, the probability of its success decreases, depending on the interdependence between the tasks.

\section{Sensors Utilized in Autonomous UAV Missions}\label{sec:sensors}

This section discusses several kinds of sensors used in UAVs and their roles in ensuring mission success. UAVs are equipped with a wide range of sensors to enable a wide range of 
tasks under various circumstances. These sensors not only improve navigation and targeting accuracy but also allow the UAV to adapt to different environments and 
operational conditions. Proper sensor configuration enhances mission success through precise navigation, target identification, engagement accuracy, 
and damage assessment \cite{chen-2011}. 
 
The authors of \cite{javaid2021} give a comprehensive evaluation of relevant sensors, while \cite{Sensorpa75} discusses more advanced sensors used specifically in UAVs 
within the military sector.

\subsection{Navigation and Stability Sensors}
\begin{itemize}
    \item \textbf{GPS Sensor}: Provides precise location data crucial for navigation and spatial orientation throughout the mission;
    \item \textbf{Accelerometer}: Measures the UAV's acceleration, aiding in flight dynamics analysis and stability during various phases of the mission;
    \item \textbf{Gyroscope}: Ensures stability in flight by maintaining angular velocity and orientation, critical for accurate targeting;
    \item \textbf{Anemometer}: Assesses wind speed and direction, feeding data to navigation systems for flight adjustments.
\end{itemize}

\subsection{Energy Monitoring Sensors}
\begin{itemize}
    \item \textbf{Battery Sensor}: Monitors battery health and charge level, ensuring sufficient power for mission completion.
\end{itemize}

\subsection{Advanced Surveillance Sensors}
\begin{itemize}
    \item \textbf{Electro-Optical Sensors}: Include high-resolution cameras and infrared sensors. During daylight, high-resolution cameras provide detailed visual data, 
while infrared sensors offer thermal imaging for night-time operations;
    \item \textbf{Synthetic Aperture Radar (SAR)}: Enables terrain analysis and change detection, effective in various weather conditions and light availability;
    \item \textbf{Multispectral and Hyperspectral Sensors}: Capture data across multiple light wavelengths, providing comprehensive environmental information;
    \item \textbf{Laser Range Finders and Laser Illuminators}: Enhance the accuracy of distance measurements and target illumination;
    \item \textbf{Gyro-Stabilized Systems}: Ensure stable imaging, crucial for surveillance and reconnaissance tasks.
\end{itemize}

\subsection{Specialized Sensor Systems}
\begin{itemize}
    \item \textbf{Specialized Imaging Systems}: Such as the AN/DVS-1 COBRA system, designed for specific military tasks like mine detection in beach surf zones.
\end{itemize}

\subsection{Sensor Networking and Data Integration}
\begin{itemize}
    \item The integration of these sensors into a networked system, like the Mini-Micro Data Link System (M2DLS), allows for the aggregation and efficient processing of data 
from various sources, enhancing the UAV's operational capabilities.
\end{itemize}

The integration of these sensors into the UAV platform not only assists in target neutralization but also ensures the safety and efficiency of the UAV throughout the mission. 
The advanced sensor suite significantly enhances the UAV's operational capabilities, enabling it to perform crucial tasks with high precision and minimal 
collateral damage \cite{chen-2011, javaid2021, olariu-2004}.

\section{Machine Learning Methodology}\label{sec:ML}

Autonomous UAVs for military tasks show potential when using ML algorithms. Although AI has not been widely used on the battlefield 
settings, the consensus among military analysts is that AI technologies (including ML) could have a significant impact on future wars \cite{hoehn-2022,konert2021}. 
Indeed, advanced ML algorithms can learn from past experience, adapt to novel conditions, and make {\em correct} decisions autonomously. All of this enhances 
the UAVs' ability to execute complex tasks and navigate challenging environments. Furthermore, the capacity of ML to process and interpret vast amounts of 
data in real-time can enhance the situational awareness of autonomous UAVs, thereby improving their precision and efficiency \cite{Suresh2019}.

We used analytical expressions to figure out how likely it is that a UAV mission will succeed in the previous section. In this section, we use an ML  
model \cite{hoehn-2022,konert2021} that can be trained on data from previous training missions, put on the UAV, and used to make decisions in real-time about 
its final mission. The UAVs' situational awareness, accuracy, and efficiency are improved because of ML's ability to manage and analyze enormous data volumes 
in real-time  \cite{Suresh2019}. In this work, we have used a Random Forest Model to identify successful UAV missions based on the provided features of 
the mission tasks.

\subsection{The Random Forest Model}

For the analysis of our UAV mission dataset, we chose the Random Forest (RF) model, a powerful ML classifier introduced by \cite{breiman2001}. This model uses a majority vote 
approach to assign a class based on the predictions of many decision trees. We selected the RF technique because of its better capability to handle high-dimensional 
data and resistance to over-fitting, which is critical given the multidimensional nature of our synthetic dataset. As indicated, the dataset includes a wide range of 
features from multiple sensors, each of which contributes complex data about the UAV's performance across various mission tasks.

Our RF model implementation relies on the Scikit-learn package \cite{pedregosa2011}, which is well-known for its efficiency and applicability in ML tasks. The 
number of decision trees in our model was set to n \= 500. This value provides an appropriate balance between computing efficiency and model accuracy. To prevent over-fitting, 
the maximum depth for each tree was limited to five levels.

The dataset was randomly divided into two parts for training and validation: 80\% for training and 20\% for testing. The strength of an RF lies in its ability to utilize the 
data discrimination capabilities of individual trees, creating an effective classification model. This feature is very useful for our dataset, which consists of $P$ data 
points and $Q$ characteristics and covers a wide range of mission-specific metrics and sensor readings \cite{jain2021}. By combining decisions from several trees, the RF model 
is able to handle the complicated and possibly non-linear relationships in our dataset. This dataset includes mission-specific metrics (like the success probability of each task) 
and multiple sensor readings (like GPS coordinates, battery levels, and environmental sensors).

\subsection{Data Description}

UAV training is critical for preparing UAVs for real-world missions. During these training sessions, UAVs are deployed to carry out simulated missions that closely match 
actual military conditions. A critical component of that process is the thorough collection and analysis of data by human experts. From takeoff to return to base, these 
experts carefully monitor and assess a wide range of parameters linked to each task of their mission. They meticulously calculate a success ratio for each mission, capturing 
the effectiveness and precision of the UAV's performance in various environments. The cumulative assessment of these task-specific success rates, together with a thorough review of 
the mission's overall execution, helps the human experts assess whether the operation may be classified as successful or unsuccessful. This detailed, expert-driven analysis 
is critical to improving UAV capabilities and ensuring their ability to prepare for real deployment.

\begin{figure}[!ht]
    \centering
    \includegraphics[width=9 cm]{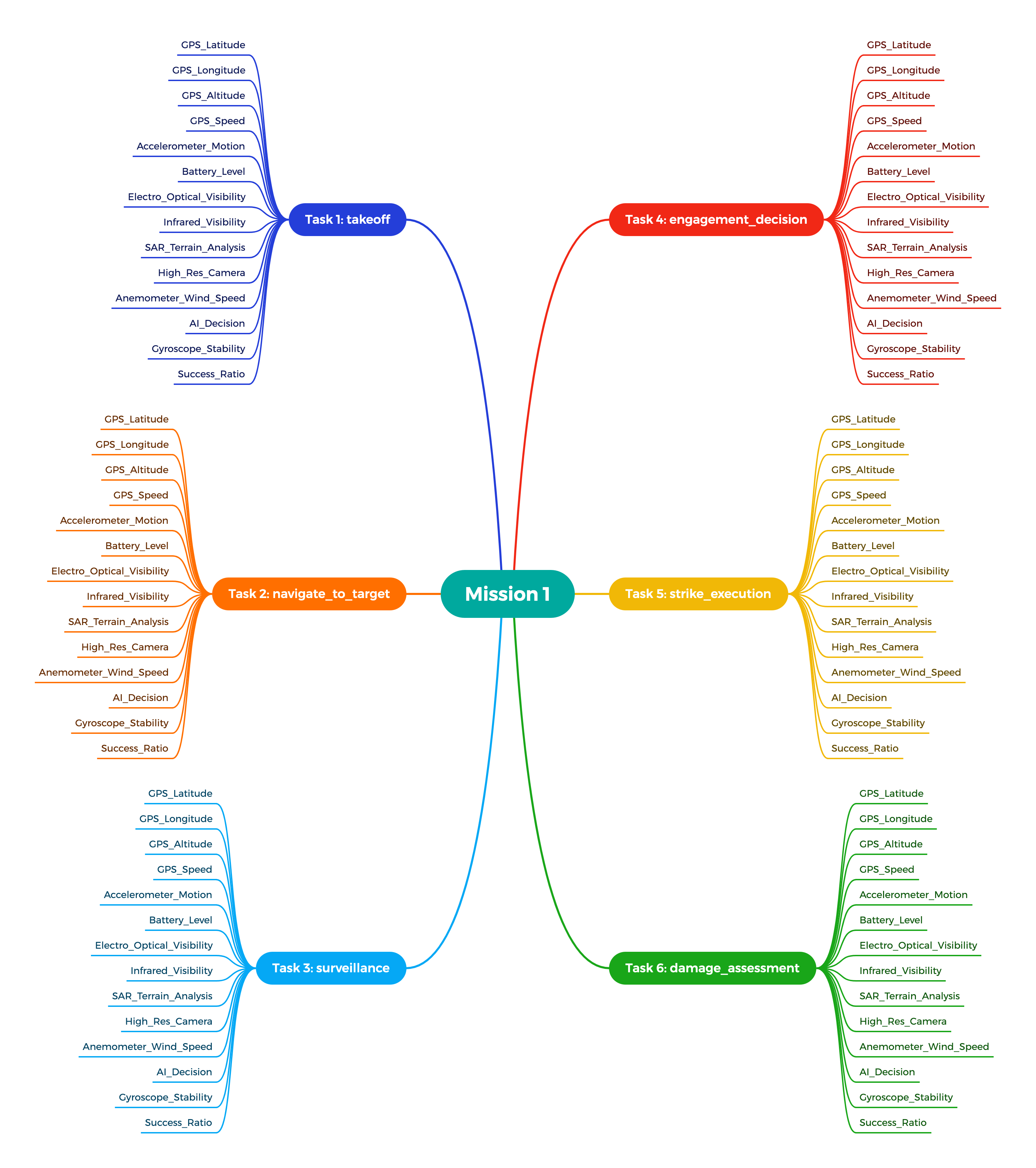}
    \caption{\em Comprehensive overview of UAV training mission data} 
    \label{fig:UAV_m}
\end{figure}

Because we did not have access to actual military UAV sensor data (which, understandably is not in the public domain), a synthetic dataset has been created to closely simulate 
these sensor readings. Figure \ref{fig:UAV_m} shows a detailed 
overview of a simulated dataset of the UAV training missions. The work addresses the lack of real-world operational data in UAV missions, particularly in sensitive operations 
such as high-value target neutralization.

The synthetic dataset aims to capture realistic mission situations with feature patterns and relationships based on actual UAV mission characteristics. It includes a variety 
of characteristics of a UAV's functioning, including task-related variables, environmental conditions, and performance metrics. For example, "GPS{\_}Latitude", "GPS{\_}Longitude",
and "GPS{\_}Altitude" offer geographical positioning, while "Battery{\_}Level" and "AI{\_}Decision" indicate the UAV's operating state and 
autonomous decision-making ability, respectively. It also shows the task success ratio, offering insight into the UAV's performance and operational efficacy 
throughout the simulated flight. 

Each mission phase, from "takeoff" to "return{\_}to{\_}base," is characterized by relevant features. These include environmental sensors like "Electro{\_}Optical{\_}Visibility" 
and "Infrared{\_}Visibility," essential for understanding the conditions under which the UAV operates. The dataset also includes decision points, such as target 
identification and engagement, based on fused sensor data and predefined rules.

The synthetic data is used as a training dataset for ML algorithms, which are designed to predict mission success and uncover important variables influencing results. 
The UAV system can learn to forecast mission outcomes and optimize its decision-making process in various settings by training the ML model on this synthetic dataset.

This dataset\footnote{The dataset and the model are provided in the following: \cite{aljohani:online}}, which was created to reflect the complexity of real-world UAV 
operations, serves as an excellent base for specifically seeking mission outcomes and measuring success indicators. It provides information about the dataset and model 
availability. This approach allows for risk-free, complete training and evaluation of UAV systems, ensuring capability for a wide range of operational scenarios and 
improving overall mission efficacy.

\subsubsection*{} 
Although creating synthetic data has advantages, it also has drawbacks. The synthetic datasets' lack of reality and accuracy is a major cause for concern. Synthetic data can 
identify patterns and connections, but it often cannot represent the complicated changes that exist in real-world data. Working with complex data types—like natural language 
text or photos, where syntactic accuracy, grammar, and visual elements are crucial—makes the issue worse. Moreover, evaluating the veracity of AI data poses an extra challenge. 
It's possible that synthetic datasets don't accurately or consistently represent the complexity and anomalies present in real data, making them unreliable as a basis for 
training models. Complicating matters further is the fact that synthetic data synthesis relies on real-world data. The synthetic data may become less helpful over time 
if the underlying data is incorrect or changes, leading to constant updates and monitoring \cite{Thebenef66:}.

There are strategies that mitigate the drawbacks of synthetic data regardless of these challenges. It is important to diversify the generated data; this can help make the 
synthetic dataset more realistic of real-world events by ensuring that the data covers a wide range of characteristics. Using solid data metrics is another important 
tactic. Metrics such as recall, accuracy, and precision are useful for assessing and improving the quality of synthetic datasets. It is necessary to conduct routine 
testing of the generated data against the features and biases of real-world data. It is possible to detect and address biases or inaccuracies in the synthetic dataset by 
using statistical tests and metrics for this purpose. Furthermore, to keep the synthetic datasets current and reflective of real-world events, it is important to monitor 
changes in the real-world data and update them accordingly. By implementing these strategies, one can increase the reliability and accuracy of synthetic data, converting it 
into a stronger tool for their data-driven tasks \cite{Thebenef66:}.

\subsubsection{UAV Operational Data Analysis}

In addition to the previously defined parameters and success ratios, the operations of the UAV are closely linked with its responses to sensory input. While the UAV performs 
its mission, each task it executes in response to sensor data is fully reported. This methodology can be likened to a "black box" approach, in which each maneuver is 
recorded, including course corrections, altitude changes, and responses to environmental factors such as wind.

Following the completion of the mission, this detailed record allows the human expert to assess the UAV's behavior in the context of the input parameters at each instant. 
For example, if the UAV effectively adjusts its altitude under difficult wind conditions, the expert would consider the task a success based on the UAV's adept 
flexibility with environmental input. A failure to adapt or an inaccurate reaction, on the other hand, would be recorded as unsuccessful, providing significant insights 
into potential areas for development in UAV programming and decision-making algorithms.

Furthermore, our approach takes into account the future employment of smart weaponry to give rapid and clear evidence of mission success, particularly in activities such as 
strike execution. Smart missiles with on-board computers can transmit real-time data and pictures as they approach and strike their target. This advanced equipment 
provides a more direct and reliable technique for confirming target impact than traditional methods such as post-mission reports or external sources such as spy 
satellites or ground operations.

However, it is important to point out that the use of smart weapons for mission success confirmation is only one component of a larger scheme. Alternative verification methods, 
such as ground reports or satellite imaging, are considered valid in scenarios where smart weapons are not employed or available.

This enhanced approach to data analysis and mission evaluation corresponds to the growing nature of UAV technology and military methods. We aim to provide an integrated 
view of UAV mission success and its drivers by combining both traditional data analysis methods and new smart weapons capabilities.

\subsection{Model Evaluation and Results}

The evaluation of classification models includes a wide range of metrics \cite{sujatha2020}, each providing distinct perspectives on the model's performance. These metrics are 
of utmost significance in determining the model's efficacy, especially in situations involving particular demands such as class imbalance or varying costs linked to various 
types of classification errors.

\subsubsection{Evaluation Metrics}
In the following, we explain the metrics for evaluating the classification models \cite{Classifi90}.\\

Accuracy is the most straightforward metric. The metric shows the ratio of accurate projections (including true positives and true negatives) to the overall count of cases analyzed.

\begin{equation}
    \text{Accuracy} = \frac{\text{Number of Correct Predictions}}{\text{Total Number of Predictions}}
\end{equation}

Precision is crucial when the cost of a false positive is large. In such cases, it is critical to reduce the rate of false positives to avoid the potentially negative implications 
of wrong positive classifications. As a result, in situations where the implications of mistaking a negative instance for a positive instance are severe, precision 
becomes a more essential measure than just improving total accuracy.

\begin{equation}
    \text{Precision} = \frac{\text{True Positives}}{\text{True Positives} + \text{False Positives}}
\end{equation}

Recall (sensitivity) is especially important when missing a positive instance (false negative) has a high cost. In such cases, identifying as many true positive occurrences 
as possible is critical, even if it results in a higher number of false positives. A strong recall is critical in situations where the consequences of missing a positive 
case are severe, exceeding the disadvantages of false positive errors.

\begin{equation}
    \text{Recall} = \frac{\text{True Positives}}{\text{True Positives} + \text{False Negatives}}
\end{equation}

The F1-score is especially useful when precision and recall must be balanced. In a classification task, for example, where both false positives and false negatives are costly, 
the F1-score gives a single statistic that balances these two characteristics. It is the harmonic mean of precision and recall, which ensures that both metrics contribute 
equally to the overall score.

\begin{equation}
    \text{F1 Score} = 2 \times \frac{\text{Precision} \times \text{Recall}}{\text{Precision} + \text{Recall}}
\end{equation}

The AUC is a common ML statistic for binary classification. A higher AUC value generally implies a better model because it demonstrates that the model can distinguish between 
positive and negative classes across all feasible thresholds. This is especially useful for analyzing models in cases where the ideal classification threshold is unknown 
and must be altered based on the specific costs or benefits associated with true positives, false positives, true negatives, and false negatives.

\subsubsection{Evaluation and results}

We assessed the performance of our model random forest using the previously described measures. Its overall efficacy can be seen by the large number of outcomes it correctly 
predicted, with an accuracy of 0.87. Its precision of 0.79 and perfect recall of 1.00, in particular, show that it can effectively detect positive outcomes while 
reducing false positives. Its F1 score of 0.88 further illustrates this performance balance.

We also compared our RF model to other classifiers; SVM (LibSVM), AdaBoost, Naive Bayes, and Bagging all show a similar pattern of accuracy, with each model obtaining an 
accuracy of 0.87. Table \ref{tab:metrics} shows the outcome of these models. RF, SVM (LibSVM), Naive Bayes, and Bagging with Decision Trees all have similar precision values 
of 0.79; however, AdaBoost has a slightly higher precision of 0.80. RF, SVM (LibSVM), Naive Bayes, and Bagging with Decision Trees all maintain a perfect recall score of 1.0, 
but AdaBoost has a slightly lower recall of 0.96. All models had similar F1 scores, indicating a fair trade-off between recall and precision. AdaBoost trails slightly 
behind with an F1 score of 0.87, while the RF model, SVM (LibSVM), Naive Bayes, and Bagging with Decision Trees also have an F1 score of 0.88. These results show that 
all classifiers function well, with just a little difference in metrics; however, the Random Forest model shows a slight edge, especially in terms of precision and recall.

\begin{table}[h]
\centering
\begin{tabular}{|l|c|c|c|c|}
\hline
\textbf{Model} & \textbf{Accuracy} & \textbf{Precision} & \textbf{Recall} & \textbf{F1 Score} \\ \hline
Random Forest & 0.87 & 0.79 & 1.00 & 0.88 \\ \hline
SVM (LibSVM) & 0.87 & 0.79 & 1.00 & 0.89 \\ \hline
AdaBoost & 0.86 & 0.80 & 0.96 & 0.87 \\ \hline
Naive Bayes & 0.87 & 0.79 & 1.00 & 0.88 \\ \hline
Bagging & 0.87 & 0.79 & 1.00 & 0.88 \\ \hline
\end{tabular}
\caption{Performance Metrics of Classification Models}
\label{tab:metrics}
\end{table}

We evaluated the effectiveness of the models for predicting the outcome of UAV missions, we noticed significant differences in their confusion matrices. The RF, SVM (LibSVM), 
Naive Bayes, and Bagging with Decision Trees models show exceptional accuracy in correctly identifying successful missions, as indicated by the absence of any false 
negatives in their confusion matrices. The RF algorithm produces a confusion matrix of [[1452, 529], [0, 2019]], the SVM (LibSVM) algorithm shows a confusion matrix 
of [[1457, 524], [0, 2019]], the Naive Bayes algorithm provides a confusion matrix of [[1445, 536], [0, 2019]], and the Bagging with Decision Trees algorithm produces a 
confusion matrix of [[1451, 530], [0, 2019]]. This shows an increased ability to identify successes in missions. Nevertheless, these models demonstrated a significant 
amount of false positives, suggesting an ability to frequently expect success.

\begin{figure*}[ht]
    \centering
    \begin{subfigure}{.2\textwidth}
        \centering
        \includegraphics[width=\linewidth]{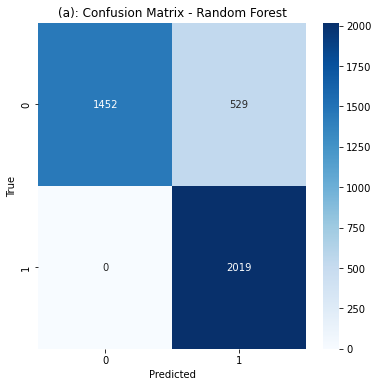}
      
    \end{subfigure}%
    \hfill
    \begin{subfigure}{.2\textwidth}
        \centering
        \includegraphics[width=\linewidth]{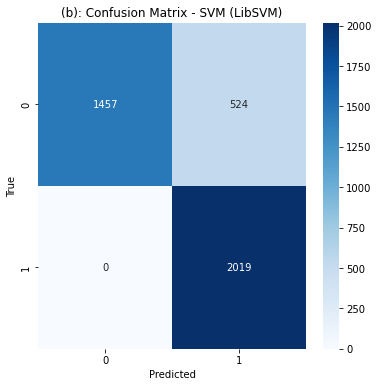}
        
    \end{subfigure}%
    \hfill
    \begin{subfigure}{.2\textwidth}
        \centering
        \includegraphics[width=\linewidth]{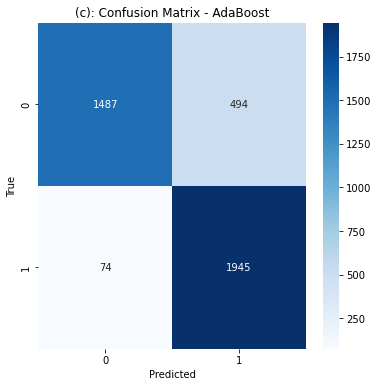}
      
    \end{subfigure}
    
    \vspace{1em}

    \hfill 
    \begin{subfigure}{.2\textwidth}
        \centering
        \includegraphics[width=\linewidth]{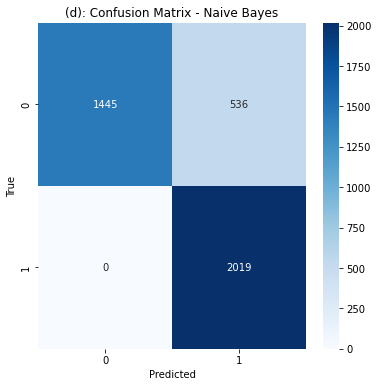}
       
    \end{subfigure}%
    \hfill 
    \begin{subfigure}{.2\textwidth}
        \centering
        \includegraphics[width=\linewidth]{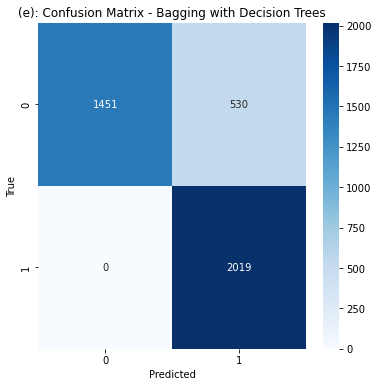}
        
    \end{subfigure}
    \hfill\phantom{\begin{subfigure}{.2\textwidth}\end{subfigure}} 
    
    \caption{Confusion matrices for five different classification models. (a) Random Forest model. (b) SVM (LibSVM) model. (c) AdaBoost model. (d) Naive Bayes model. (e) Bagging with Decision Trees model.}
    \label{fig:Conf}
\end{figure*}

On the other hand, the AdaBoost model shows an equal performance in detecting both successful and unsuccessful missions, as evidenced by its confusion matrix 
of [[1487, 494], [74, 1945]]. Although many cases of false negatives, this model had a reduced rate of false positives in comparison to the other models. This balance 
indicates a more precise ability of the mission parameters. Figure \ref{fig:Conf} shows the confusion matrices for the five different classification models.

\begin{figure*}[ht]
    \centering
    \begin{subfigure}{0.22\textwidth}
        \centering
        \includegraphics[width=\linewidth]{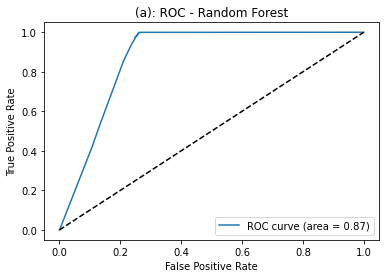}
        \caption{Random Forest}
    \end{subfigure}%
    \begin{subfigure}{0.22\textwidth}
        \centering
        \includegraphics[width=\linewidth]{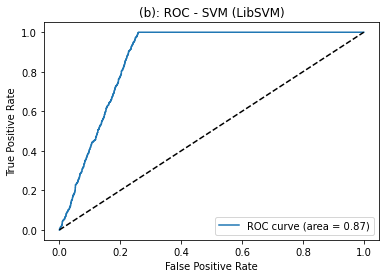}
        \caption{SVM (LibSVM)}
    \end{subfigure}%
    \begin{subfigure}{0.22\textwidth}
        \centering
        \includegraphics[width=\linewidth]{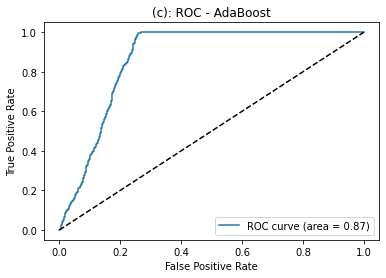}
        \caption{AdaBoost}
    \end{subfigure}%
    \begin{subfigure}{0.22\textwidth}
        \centering
        \includegraphics[width=\linewidth]{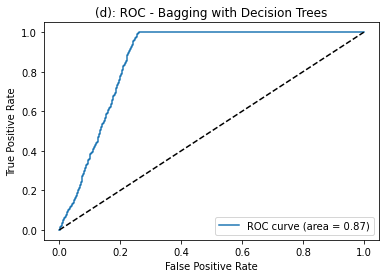}
        \caption{Bagging with DT}
    \end{subfigure}%
    
    \caption{Receiver Operating Characteristic (ROC) curves for four different classification models.}
    \label{fig:ROC}
\end{figure*}

We also evaluated our models using the Receiver Operating Characteristic (ROC) curve, which is an essential element in the evaluation of classification models since it shows how 
well the model can differentiate between classes. Figure \ref{fig:ROC} shows the ROC curves for the RF, SVM (LibSVM), AdaBoost, and Bagging with Decision Trees models. 
The models RF, SVM (LibSVM), AdaBoost, and Bagging with Decision Trees all showed ROC AUC (Area Under the Curve) values of 0.87, which is a clear indication of their high 
degree of classification effectiveness and ability to distinguish between positive and negative classifications. The Naive Bayes model was excluded from the ROC AUC 
analysis as our implementation didn't work to provide the probability estimates required for ROC curve creation.\\

\begin{figure}[ht]
    \centering
    \includegraphics[width=0.4\textwidth]{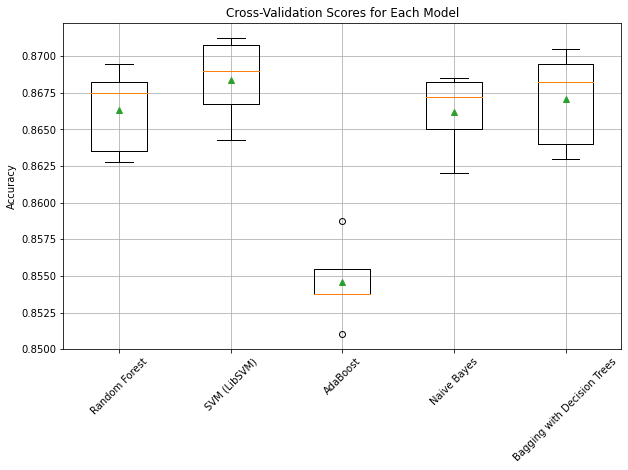}
    \caption{ Box Plot of Classification Model Cross-Validation Results. The distribution of accuracy scores from five-fold cross-validation for five different classification models is shown in this figure: AdaBoost, Random Forest, SVM (LibSVM), Naive Bayes, and Decision Tree-Based Bagging.}
    \label{fig:box}
\end{figure}

We performed a thorough cross-validation evaluation to assess the efficacy of five distinct classification models: RF, SVM (LibSVM), AdaBoost, Naive Bayes, and Bagging 
with Decision Trees. To provide accurate and trustworthy assessments, the cross-validation procedure was performed for five different runs. We created a box 
plot \ref{fig:box} by combining the data from each of the five runs to contrast the performance of these models. This box plot clearly illustrates the distribution of 
accuracy ratings for each model while also illuminating the robustness and variability of the various models.

The box plot illustrates the cross-validation results, which are as follows: The RF model demonstrated consistently high performance across the runs, as shown by its mean 
accuracy scores, which varied from 0.8555 to 0.8695. The accuracy ratings of the SVM (LibSVM) model ranged from 0.8538 to 0.8635, suggesting a consistent and similar level 
of performance. The scores for the AdaBoost model ranged from 0.8538 to 0.8708, demonstrating its efficacy across multiple iterations. The Naive Bayes model demonstrated 
competitive prediction skills with scores between 0.8588 and 0.8712. Finally, the Bagging with Decision Trees model produced scores ranging from 0.8658 to 0.8510, 
indicating that it is a reliable classifier.\\

\section{Discussion}\label{sec:discussion}

Despite the existing use of RF models for mission success prediction, integrating advanced ML techniques, particularly Deep Q-Networks, and specialized ML algorithms for 
UAV communications have the potential to further transform military UAV operations. Deep Q-Networks, as described in \cite{li2019} provides a comprehensive approach to 
real-time on-board decision-making. This technology has the potential to dramatically improve the efficiency of UAVs in power management and data collecting, which is 
critical for long-term, complex military operations. Furthermore, the authors of \cite{ben2022} emphasize the promise of ML for optimizing UAV communication systems. 
Improved communication protocols allowed by ML would not only afford better data transfer and processing but would also improve UAV fleet situational awareness and 
coordination. These advanced ML applications fit with the need for adaptability and efficiency in military scenarios, implying that AI will 
play an important role in the future of warfare. 

Furthermore, the integration of multiple and advanced ML models, such as Convolutional Neural Networks (CNNs) for image analysis, Recurrent Neural Networks (RNNs), and 
transformers for sequential data interpretation for structured mission data, provides a variety of techniques for further improving UAV capabilities. CNNs, in 
particular, can revolutionize target identification and surveillance through advanced image processing \cite{egli2020}. RNNs and transformers offer unparalleled advantages 
in analyzing temporal data, essential for real-time decision-making and strategy formulation \cite{kurunathan2023}. An integrated application that utilizes 
multiple ML models not only enhances the particular capabilities of each technique but also meets the various issues encountered in modern military operations. 
The incorporation of advanced ML algorithms represents a significant advancement in military UAV operations \cite{Artifici60}, resulting in increased operational 
effectiveness, adaptability, and strategic superiority in complex and dynamic combat scenarios.

\section{Concluding remarks}\label{sec:concl}
In this paper, we have identified the challenges related to enabling autonomous UAVs to carry out strike missions against high-value terrorists. 
We have suggested that recent developments in ledger technology, smart contracts, and ML
enable autonomous UAVs to complete these missions successfully. We have derived analytical expressions for the success of a mission 
depending on the interdependence of tasks within the mission. Last but not least, we have demonstrated an ML framework for autonomous UAVs.

Several issues are still open and are getting attention. Graph Machine Learning (GML) is a recently proposed ML technology where the power of graph representation
of information is harnessed to advantage \cite{wei-2023}. In this context, graph decomposition techniques \cite{jamison-1995} are useful tools intended to enhance the 
scalability of GML techniques.

Many other avenues for future investigations are open. One of them is mission security. While we have developed our mission with minimal communication
requirements, in the future communications may play an important role \cite{abdelsalam-2012,rawat-2014}, and ensuring a high level of security will turn out
to be essential. Yet another open problem is the type of communications and local processing that the UAV must perform \cite{nakano-1999} while in flight.

\bibliographystyle{IEEEtran}


\end{document}